\documentclass[runningheads]{llncs}
\usepackage[T1]{fontenc}

\usepackage[numbers]{natbib}

\usepackage{pdflscape}
\usepackage{graphicx, tikz}
\usetikzlibrary{positioning, fit, arrows.meta}

\usepackage{amsmath,amssymb}

\DeclareMathOperator*{\argmin}{arg\,min}
\newcommand{\bs}[1]{\boldsymbol{#1}}
\newcommand{\mb}[1]{\mathbf{#1}}
\newcommand{\mc}[1]{\mathcal{#1}}
\newcommand{\bphi}{\bar{\bs{\phi}}}

\newcommand{\bX}{\mb{X}}
\newcommand{\bz}{\mb{z}}
\newcommand{\bc}{\mb{c}}
\newcommand{\bu}{\mb{u}}

\usepackage{hyperref,cleveref}

\usepackage{todonotes}

\begin{document}
\title{Contextual Preference Distribution Learning}
%
%

\author{Benjamin Hudson\inst{1,2}\orcidID{0000-0002-9461-5895} \and
Laurent Charlin\inst{1,2,3}\orcidID{0000-0002-6545-9459} \and
Emma Frejinger\inst{1,2}\orcidID{0000-0003-1930-607X}}

\authorrunning{B. Hudson et al.}
%
\institute{Mila -- Quebec Artificial Intelligence Institute, Montreal, Quebec, Canada\\\email{ben.hudson@mila.quebec}\and
D\'{e}partement d'informatique et de recherche op\'{e}rationnelle (DIRO), Universit\'{e} de Montr\'{e}al, Montreal, Quebec, Canada \and
HEC Montr\'{e}al, Montreal, Quebec, Canada}

\maketitle              
\begin{abstract}
Decision-making problems often feature uncertainty stemming from heterogeneous and context-dependent human preferences.
To address this, we propose a sequential learning-and-optimization pipeline to learn preference distributions and leverage them to solve downstream problems, for example risk-averse formulations.
We focus on human choice settings that can be formulated as (integer) linear programs. In such settings, existing inverse optimization and choice modelling methods infer preferences from observed choices but typically produce point estimates or fail to capture contextual shifts, making them unsuitable for risk-averse decision-making.
Using a bounded-variance score function gradient estimator,
we train a predictive model mapping contextual features to a rich class of parameterizable distributions. This approach yields a maximum likelihood estimate.
The model generates scenarios for unseen contexts in the subsequent optimization phase.  
In a synthetic ridesharing environment, our approach reduces average post-decision surprise by up to 114$\times$ compared to a risk-neutral approach with perfect predictions and up to 25$\times$ compared to leading risk-averse baselines.
\end{abstract}

\section{Introduction}

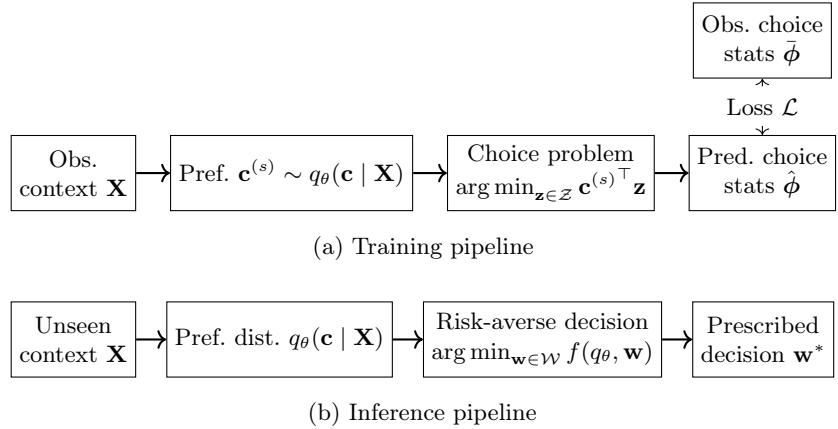
\begin{figure}
    \centering
    \begin{tikzpicture}[
      node distance = 0.4cm,
      every node/.style = {align=center},
      myNode/.style = {rectangle, draw, minimum size=1cm}
    ]
    
    \node[myNode] (x) {Obs.\\context $\mb{X}$};
    \node[myNode, right=0.45cm of x] (c) {Pref. $\mb{c}^{(s)} \sim q_\theta(\mb{c} \mid \mb{X})$};
    \node[myNode, right=0.45cm of c] (opt) {Choice problem\\$\argmin_{\mb{z}\in\mathcal{Z}} {\mb{c}^{(s)}}^\top \mb{z}$};
    \node[myNode, right=0.45cm of opt] (zpred) {Pred. choice\\stats $\hat{\bs{\phi}}$};
    \node[myNode, above=0.75cm of zpred] (zexp) {Obs. choice\\stats $\bar{\bs{\phi}}$};
    \node[fit=(x) (c) (opt) (zpred)] (training_pipeline) {};
    \node[below=0.1cm of training_pipeline] {(a) Training pipeline};
    
    \draw[->, thick] (x) -- (c);
    \draw[->, thick] (c) -- (opt);
    \draw[->, thick] (opt) -- (zpred);
    \draw[dashed, <->] (zpred) -- node[midway,fill=white]{Loss $\mc{L}$} (zexp);

    \node[myNode, below=1.2cm of x] (unseen_x) {Unseen\\context $\mb{X}$};
    \node[myNode, right=of unseen_x] (q) {Pref. dist. $q_\theta(\mb{c} \mid \mb{X})$};
    \node[myNode, right=of q] (ra_opt) {Risk-averse decision\\$\argmin_{\mb{w}\in\mathcal{W}} f(q_\theta,\mb{w})$};
    \node[myNode, right=of ra_opt] (w) {Prescribed\\decision $\mb{w}^*$};
    \node[fit=(unseen_x) (q) (ra_opt) (w)] (inference_pipeline) {};
    \node[below=0.1cm of inference_pipeline] {(b) Inference pipeline};
    
    \draw[->, thick] (unseen_x) -- (q);
    \draw[->, thick] (q) -- (ra_opt);
    \draw[->, thick] (ra_opt) -- (w);

\end{tikzpicture}
    \caption{\textbf{Our sequential learning-and-optimization approach. (a)} During training, we learn a model mapping contextual features to preference (objective function coefficient) distributions in the choice problem (an ILP). \textbf{(b)} At inference, we leverage the model to generate scenarios in a risk-averse estimate-then-optimize problem, allowing us to minimize risk introduced by uncertain human preferences.}
    \label{fig:pipelines}
\end{figure}

Human preferences vary across individuals and contexts~\cite{burton_systematic_2020}, often causing \textit{post-decision surprise}~\cite{harrison_decision_1984} in decision-making problems that depend on human choices.
We take driver-rider assignment in ridesharing as an illustrative example: while platforms recommend routes using real-time traffic~\cite{nguyen_eta_2015}, drivers often choose alternate routes to optimize their own objectives, based on personal preferences or tacit knowledge of the road network~\cite{merchan_2021_2024}.
These discrepancies introduce uncertainty into downstream problems; for example, if a driver takes an alternate route, their realized arrival time may differ from the platform's estimate, leading to the perception of unreliability. To address this issue, we propose a sequential learning-and-optimization approach (illustrated in Fig.~\ref{fig:pipelines}): we first predict human preference distributions given some contextual features (i.e.\ covariates), and then leverage them to explicitly minimize the resulting downstream risk. Ultimately, we aim to reduce post-decision surprise in decision-making problems where uncertainty stems from human choices.

Existing approaches modelling human choice behaviour with mathematical programs, such as inverse optimization~\cite{besbes_contextual_2025,chan_inverse_2025,mohajerin_esfahani_data-driven_2018,zattoni_scroccaro_learning_2024} and perturbed utility models~\cite{allen_identification_2019,fosgerau_perturbed_2022,mcfadden_theory_2012}, are typically limited by predicting point estimates rather than distributions or by failing to account for how preferences shift with context.
This renders them unsuitable for risk-averse decision-making: distributional information is necessary to quantify risk, while conditioning on context helps to avoid overly conservative decisions.

\subsubsection{Contribution statement.}
Our contributions are summarized as follows:
\begin{enumerate}
    \item We propose a method to learn context-dependent preference (i.e.\ objective function coefficient) distributions in integer linear programs (ILPs) from paired context and aggregated solution statistic observations. This is ideal when directly observing the uncertain quantities is impossible or undesirable.
    \item Our approach yields a maximum likelihood estimate (MLE), inheriting the desirable statistical properties of consistency and efficiency.
    \item Our score function gradient estimator has bounded variance and did not require post hoc variance reduction in our experiments.
    \item We demonstrate our approach as part of a sequential learning-and-optimization pipeline for decision-making under human preference uncertainty.
    In a synthetic ridesharing environment, it reduces average post-decision surprise by 2.4 -- 114$\times$ compared to a risk-neutral approach with perfect predictions and by 1.6 -- 25$\times$ compared to leading risk-averse baselines.
\end{enumerate}

\section{Related Work}\label{sec:related_work}
We frame our work as inverse optimization (IO), as we recover parameters of an optimization formulation from observed solutions. It relates closely to perturbed utility models (PUMs) and inverse reinforcement learning (IRL), which model choice behaviour as optimization problems or Markov decision processes (MDPs). 
We leverage integrated learning-and-optimization (ILO) techniques during training and deploy our model in a sequential learning-and-optimization (SLO) pipeline~\cite{sadana_survey_2025}.

\subsubsection{Inverse optimization.}
Recent research in IO addresses settings where observations are corrupted by noise or bounded rationality~\cite{chan_inverse_2025,mohajerin_esfahani_data-driven_2018} yet the underlying problem formulation is deterministic. The learning task is to recover parameters of this formulation. Contextual IO addresses settings in which these vary via known mappings~\cite{besbes_contextual_2025,zattoni_scroccaro_learning_2024}. In contrast, we tackle a setting where the latent parameters are random and the feature mapping function is unknown. The closest IO approach to ours is that of Lin et al.~\cite{lin_conformal_2024}, who recover uncertainty sets over problem parameters, but do not consider a contextual setting.

\subsubsection{Perturbed utility models.}
PUMs generalize the classical discrete choice setting~\cite{allen_identification_2019,mcfadden_econometric_1981,train_discrete_2009} and specialized variants like bundle~\cite{birge_learning_2022,chen_learning_2022,mcfadden_theory_2012} and route choice~\cite{fosgerau_link_2013,fosgerau_perturbed_2022}. These models express choice probabilities as solutions to convex, constrained optimization problems. Most current approaches model a single individual representative of the population~\cite{allen_identification_2019}: that is, they learn a function mapping from observed features to a point prediction of expected utility (negative cost). Birge et al.~\cite{birge_learning_2022} explore learning Dirac-Uniform and von Mises-distributed utilities as special cases.
A distributional approach enables simulating choices in unseen contexts, something we explore in our experiments.

\subsubsection{Inverse reinforcement learning.}
In contrast to IO and PUMs, IRL frameworks model complex choice problems as MDPs, aiming to recover their parameters from observed decisions~\cite{canoy_learn_2023,canoy_vehicle_2019,kristensen_inverse_2021,zhao_deep_2023,ziebart_maximum_2008}. The applicability of IRL heavily relies on whether a problem fits the MDP paradigm. In contrast, our approach accepts any ILP formulation and is agnostic to problem substructure and the underlying solution method.

\subsubsection{Integrated learning-and-optimization.} ILO, or decision-focused learning (DFL)~\cite{mandi_decision-focused_2024}, trains a model by considering the impact of its predictions on downstream decisions (typically measured by regret). Our training pipeline can be seen as an optimal action imitation ILO task, where the goal is to reproduce observed actions given context~\cite{sadana_survey_2025}. While many methods demonstrated on regret minimization can be readily applied to action imitation~\cite{berthet_learning_2020,minervini_adaptive_2023,niepert_implicit_2021,sun_maximum_2023,sahoo_gradient_2022,vlastelica_differentiation_2020}, we contribute a novel approach for back-propagating through ILPs using a bounded-variance score function gradient estimator.

\subsubsection{Sequential learning-and-optimization.}
We take an SLO approach, using predictions from our trained model in a distinct optimization problem. While estimate-then-optimize (ETO)~\cite{bertsimas_dynamic_2023,deng_predictive_2022,elmachtoub_estimate-then-optimize_2025,hannah_nonparametric_2010,kannan_residuals-based_2024,srivastava_data-driven_2021} and conditional robust optimization (CRO)~\cite{chenreddy_end--end_2024,chenreddy_data-driven_2022,ohmori_predictive_2021,patel_conformal_2023,persak_contextual_2023,sun_predict-then-calibrate_2023} predict conditional distributions and uncertainty sets for risk-averse decision-making, both rely on direct observations of the uncertain parameters. In contrast, our approach requires only aggregated solutions to a related problem (the choice problem), making it ideal when direct observation is impossible or undesirable, e.g.,\ for privacy concerns.

\section{Contextual Preference Distribution Learning}
In this section, we introduce Contextual Preference Distribution Learning (CPDL), demonstrate that it yields an MLE of the model parameters, and discuss the variance of its stochastic gradient estimator.
\subsection{Learning problem formulation}
We model human choice behaviour as an ILP with binary variables (the ``choice problem''). This formulation is compact, yet can represent a wide variety of choice settings where an individual selects from a discrete set of alternatives, such as shortest paths~\cite{berthet_learning_2020,fosgerau_perturbed_2022}, $k$-subset selection~\cite{birge_learning_2022,mcfadden_theory_2012,niepert_implicit_2021}, and discrete choice~\cite{allen_identification_2019,mcfadden_econometric_1981,train_discrete_2009}.
Given a vector of preferences $\bc$, the choice problem yields optimal solutions
\[
    \mc{Z}^*(\bc,\bu):=\argmin_\bz\ \{\bc^\top\bz\ \vert\ \bz \in \mc{Z}(\bu), \bz \in \{0,1\}^m\},
\]
where $\mc{Z}(\mb{u})$ defines the feasible region given some exogenous variables $\mb{u}$.
We assume an individual selects an optimal solution uniformly at random, denoted by $\bz^* \in_R \mc{Z}^*(\bc,\bu)$.
This defines the conditional choice distribution $p(\bz|\bc,\bu)$ as having equal mass over all elements in the optimal set. To reduce notational clutter, we omit conditioning on $\bu$ where the context is unambiguous.
Human preferences are varied, context-dependent, and can be represented by a distribution $p(\bc|\bX)$, where $\bX$ is a $m\times n$ contextual feature matrix, containing an $n$-dimensional feature vector for each element of $\bc$.
Thus, we can express the population choice distribution as
\[
p(\bz|\bX)=\int p(\bz|\bc)p(\bc|\bX)d\bc.
\]

We now describe the learning problem. Our goal is to recover the distribution $p(\bc|\bX)$ in a way that can generalize to previously unobserved contexts. We leverage amortized inference to approximate this distribution using a parameterized model $q_\theta(\bc|\bX)$ shared across instances.
We model the target and predicted choice distributions $p(\bz|\bX)$ and $q_\theta(\bz|\bX)=\int p(\bz|\bc)q_\theta(\bc|\bX)d\bc$ as discrete exponential-family distributions, allowing us to obtain an MLE of the model parameters $\theta$ by optimizing them subject to a moment matching condition~\cite{murphy_probabilistic_2023,niepert_implicit_2021,ziebart_maximum_2008}. For our model, this condition is given by
\[
\mathbb{E}_{p(\bz|\bX)}[\phi(\bz)] - \mathbb{E}_{q_\theta(\bc|\bX)}\left[\mathbb{E}_{p(\bz|\bc)}[\phi(\bz)]\right] = 0,
\]
where $\phi(\bz)$ are sufficient statistics of the distribution of $\bz$ (target or predicted).
We define the inverse problem as
\[
\begin{aligned}
    \theta^* \in \argmin_{\theta \in \Theta} \left\|\mathbb{E}_{p(\bz|\bX)}[\phi(\bz)] - \mathbb{E}_{q_\theta(\bc|\bX)}\left[\mathbb{E}_{p(\bz|\bc)}[\phi(\bz)]\right]\right\|_2^2.
\end{aligned}
\]
Unlike previous work~\cite{niepert_implicit_2021}, we optimize this objective directly using stochastic gradient descent.

\subsection{Solution approach}
Let $\bX$ be an observed feature matrix and $\bphi:=\mathbb{E}_{p(\bz|\bX)}[\phi(\bz)]$ be the sufficient statistics of the observed choice distribution. Let the expected sufficient statistics under the predicted distribution be $\hat{\bs{\phi}}:=\mathbb{E}_{q_\theta(\bc|\bX)}\left[\mathbb{E}_{p(\bz|\bc)}[\phi(\bz)]\right]$.
In this paper, we define $\phi(\bz)=\bz$, corresponding to first-moment matching, although we experiment with matching higher-order moments in App.~\ref{app:higher_moments}.
The loss of the inverse problem and its gradient are given by
\begin{align}
    \mc{L}(\bX,\bphi,\theta)&=\left\|\bphi - \hat{\bs{\phi}}\right\|_2^2\label{eqn:loss},\\
    \nabla_\theta \mc{L}(\bX,\bphi,\theta)&=\left(\bphi - \hat{\bs{\phi}}\right)\nabla_\theta \mathbb{E}_{q_\theta(\bc|\bX)}\left[\mathbb{E}_{p(\bz|\bc)}[\phi(\bz)]\right]\nonumber\\
    &=\left(\bphi - \hat{\bs{\phi}}\right)\mathbb{E}_{q_\theta(\bc|\bX)}\left[\mathbb{E}_{p(\bz|\bc)}[\phi(\bz)]\nabla_\theta \log q_\theta(\bc|\bX)\right]\label{eqn:loss_grad}.
\end{align}

We approximate $\hat{\bs{\phi}}$ with the Monte Carlo estimate:
\[
\hat{\bs{\phi}}\approx \frac{1}{KL}\sum_{k=1}^K\sum_{l=1}^L\phi\left({\bz^*}^{(kl)}\right)\quad \mathrm{where}\ {\bz^*}^{(kl)}\in_R \mc{Z}^*(\bc^{(k)}, \bu)\ \mathrm{and}\ \bc^{(k)}\sim q_\theta(\bc|\bX).
\]
The hyperparameters $K$ and $L$ are the number of samples drawn from the learned cost distribution and the conditional choice distribution, respectively.
We find $L=1$ is sufficient to approximate $\mathbb{E}_{p(\bz|\bc)}[\phi(\bz)]$, as the choice problem typically has a single solution for a given cost vector.
While this estimate involves solving the choice problem $KL$ times, the instances in our setting are not computationally challenging. Computation can be accelerated with warm-starting, solution caching~\cite{mulamba_contrastive_2021}, or GPU parallelization~\cite{lu_mpax_2025}.
We compute a stochastic gradient estimate by re-weighting each solution by the score of the generating sample from $q_\theta(\bc|\bX)$:
\[
\nabla_\theta \mc{L}(\bX,\bphi,\theta)\approx\left(\bphi - \hat{\bs{\phi}}\right)\left(\frac{1}{K}\sum_{k=1}^K\left(\frac{1}{L}\sum_{l=1}^L \phi\left({\bz^*}^{(kl)}\right)\right)\nabla_\theta \log q_\theta (\bc^{(k)}|\bX)\right).
\]

While the solution distribution is restricted to the discrete exponential family, $q_\theta(\bc|\bX)$ can be any parametrizable distribution with a tractable score function (e.g., normalizing flows).
Despite being unconstrained, our experiments show we recover the ground-truth distributional parameters up to an invariant transformation when the distribution is correctly specified.

\subsubsection{Variance of the gradient estimator.}\label{sec:variance}

Mohamed et al.~\cite{mohamed_monte_2020} show the variance of a score function gradient estimate of a function $f$ under a smoothing distribution $p_\theta(x)$ is given by
\[
\mathrm{Var}(\hat{\nabla}_\theta f)=\mathbb{E}_{p_\theta(x)}\left[(f(x)\nabla_\theta\log p_\theta(x))^2\right] - (\hat{\nabla}_\theta f)^2.
\]
Large magnitudes of $f$ cause this variance to explode, leading to training instability. In our approach, $f:=\mathbb{E}_{p(\bz|\bc)}[\phi(\bz)]$ represents the moments of the choice distribution. Because these are bounded in $[0, 1]$, the variance of our gradient estimate is bounded from above by $\mathbb{E}_{p_\theta(x)}\left[(\nabla_\theta\log p_\theta(x))^2\right]$,
which corresponds to the diagonal elements of the Fisher information matrix.
Empirically, this allows CPDL to use higher learning rates than methods with unbounded $f$ (e.g.,\ REINFORCE~\cite{silvestri_score_2024}) without post hoc variance reduction. We compare CPDL and REINFORCE in detail in App.~\ref{app:reinforce_comparison}.

\section{Numerical Experiments}
We benchmark CPDL against other methods capable of generating scenarios in a synthetic ridesharing environment.

\subsection{Experimental setting}
Our environment comprises a synthetic data-generating process representing driver route choice for the ``estimate'' phase and a risk-averse driver-rider assignment problem for the ``optimize'' phase.

\subsubsection{Route choice problem.}
Our data-generating process is similar to previous work ~\cite{berthet_learning_2020,lin_conformal_2024,niepert_implicit_2021,vlastelica_differentiation_2020}, but features stochastic edge costs whose distributions are feature-dependent. We generate a set of choices (the \textit{samples} $\mc{S}$) for each context (the \textit{instance}, $i \in \mc{I}$). This is standard in settings with categorical features, where multiple observations per category are expected~\cite{canoy_learn_2023,guo_context-aware_2020}.

We begin by uniformly sampling two global $n$-dimensional vectors $\bs{\beta}_\mu$ and $\bs{\beta}_\sigma$.
We generate instances representing grid graphs, each having a $|\mc{E}|\times n$ contextual feature matrix $\bX_{i}$ sampled from a standard normal distribution and edge cost distributions $\mathrm{LogNormal}(\bX_i\bs{\beta}_\mu,\bX_i\bs{\beta}_\sigma)$. We use a log-normal distribution for its positive support and use in travel-time estimation literature~\cite{elmasri_predictive_2023,hunter_path_2009,westgate_large-network_2016}. 
For each instance, we generate samples by drawing edge costs and solving the resulting shortest path problems $\{{\bz^*}^{(s)}_i|\ \forall s\in \mc{S}\}$.
We define $\phi(\bz)=\bz$, corresponding to matching the first choice distribution moment: the expected edge selection probability $\bar{\mb{p}}_i=\frac{1}{|\mc{S}|}\sum_{\mc{S}}{\bz^*}_i^{(s)}$.
The learner observes only $\bX_i$ and $\bar{\mb{p}}_i$. The expression of the data-generating process is shown in App.~\ref{app:dgp}.

\subsubsection{Risk-averse assignment problem.}
We leverage the trained model to solve a risk-averse variant of the driver-rider assignment problem, directly addressing risk introduced by uncertain human preferences.
Given unseen features $\bX$, we sample preferences from the predicted distribution $q_\theta(\bc|\bX)$. For each scenario $k\in \mc{K}$, we determine the perceived-shortest paths from all drivers to all riders and compute their cost according to a decision cost vector $\mb{g}$ --- the first column of $\bX$ in our experiments. We assemble these into a cost matrix $\mb{G}^{(k)}$, representing a scenario. Finally, we obtain an assignment $\mb{W}^*$ by solving the conditional value-at-risk (CVaR) minimization problem formulation in App. \ref{app:ra_assignment} with $\alpha=0.95$.

\subsection{Evaluation metrics}
Perception of unreliability is caused by both early and late arrivals. Thus, we define post-decision surprise as the squared difference between the expected decision cost under the ground-truth distribution $p(\bc|\bX)$ and the predicted distribution $q_\theta(\bc|\bX)$. Mathematically,
\begin{equation}\label{eqn:post-decision_surprise}
\mc{L}_{PDS}(\mb{W}^*,q_\theta)=\left(\mathbb{E}_{\bs{\xi}\sim p(\bc|\bX)}\left[g(\mb{W}^*,\bs{\xi})\right] - \mathbb{E}_{\bs{\zeta}\sim q_\theta(\bc|\bX)}\left[g(\mb{W}^*,\bs{\zeta})\right]\right)^2,
\end{equation}
where $g$ is the decision cost in a given scenario (i.e. given a realization of uncertainty $\bs{\xi}$ or $\bs{\zeta}$). We approximate both expectations with Monte Carlo estimates (App.~\ref{app:ra_assignment}).
Because our risk-averse assignment formulation effectively penalizes only late arrivals, we also report this one-sided post-decision \textit{disappointment} in App.~\ref{app:disappointment}. Finally, to measure the models' capacity to learn the underlying preference distributions, we report the test loss and the squared correlation coefficient ($R^2$) between the predicted and ground-truth parameters (App.~\ref{app:dist_recovery}).

\subsection{Baselines}
We compare against four baselines mapping context to distributions:
DPO~\cite{berthet_learning_2020}, AIMLE~\cite{minervini_adaptive_2023}, REINFORCE~\cite{silvestri_score_2024}, and MaxEnt IRL~\cite{ziebart_maximum_2008}. We omit IO and PUM~\cite{fosgerau_perturbed_2022} baselines because they make point predictions, and do not allow sampling scenarios. While all baselines share the same encoder (an MLP with two 32-dimensional hidden layers) and loss \eqref{eqn:loss}, they estimate gradients through the choice problem differently. The encoder parameters are shared across edges.
Additionally, we evaluate two baselines with perfect predictions: GTD-RN solves a risk-neutral assignment using ground-truth expected preferences, while GTD-RA solves the risk-averse assignment with the ground-truth preference distributions. Further baseline details are in App.~\ref{app:baselines}.

\subsection{Results}
\begin{figure}
    \centering
    \includegraphics[width=0.49\linewidth]{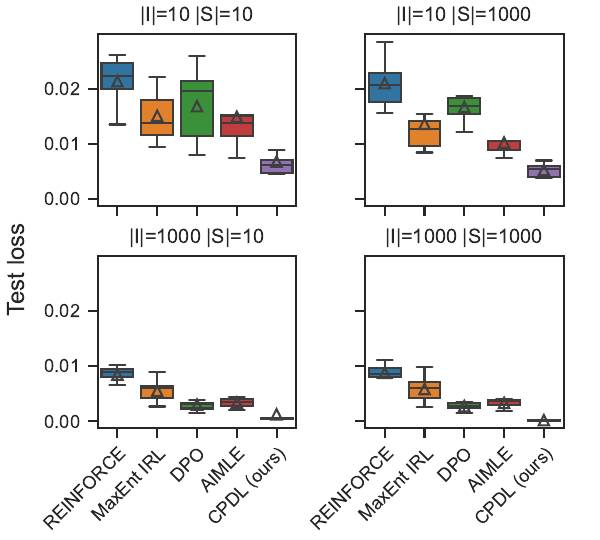}
    \includegraphics[width=0.49\linewidth]{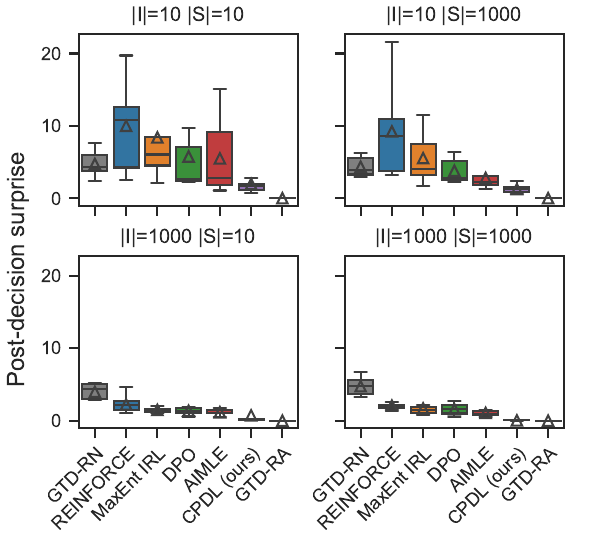}
    \caption{
    \textbf{Model performance across dataset scale and fidelity.} Test loss (left) and post-decision surprise (right) as a function of training set size (rows) and observation fidelity (columns), for nine random seeds. Increasing the number of instances ($|\mc{I}|$) helps the model learn the context-to-distribution mapping, while increasing the number of samples per instance ($|\mc{S}|$) reduces noise in the choice distribution statistics ($\bphi$).
    }
    
    \label{fig:main_results}
\end{figure}

Fig.~\ref{fig:main_results} demonstrates that our method reconstructs the choice distribution in unseen contexts with the highest accuracy (measured by test loss) and generates realistic scenarios, yielding prescribed assignments with low post-decision surprise.
Interestingly, while baselines can produce assignments with post-decision high surprise, they rarely lead to post-decision disappointment (Fig.~\ref{fig:main_results_app}); we compare these metrics and discuss this phenomenon further in App.~\ref{app:disappointment}. Finally, App.~\ref{app:dist_recovery} details the models' ability to recover the underlying preference distributions, and App.~\ref{app:higher_moments} explores the effects of matching higher-order moments.


\section{Conclusion}
We propose a method to learn contextual preference distributions from observed contexts and aggregated solution statistics. By leveraging a bounded-variance stochastic gradient estimator, our approach yields an MLE of the model parameters. We demonstrate its ability to generate realistic scenarios for unseen contexts in a synthetic ridesharing environment, where we prescribe driver-rider assignments accounting for the uncertainty stemming from human choices.
Our approach reduces average post-decision surprise by 2.4--114$\times$ compared to a risk-neutral approach with perfect predictions, and by 1.6--25$\times$ compared to leading risk-averse baselines. Future work will explore normalizing flows and broader choice settings, like discrete choice and tour choice (VRP) problems.

\bibliographystyle{splncs04}
\bibliography{references}
\clearpage
\appendix
\section{Additional theoretical results}
\subsection{Comparison to REINFORCE}\label{app:reinforce_comparison}
CPDL computes the loss \eqref{eqn:loss} for each datapoint using a batch of samples from the learned distribution $q_\theta(\bc|\bX)$. REINFORCE does the opposite: it computes the loss for a batch of datapoints for each sample from the learned distribution.
This approach effectively treats the entire distribution-to-loss pipeline as a black box and yields a different gradient from \eqref{eqn:loss_grad}~\cite{silvestri_score_2024}.
Mathematically,
\[
\begin{aligned}
\mathcal{L}_{\mathrm{REINFORCE}}(\bX,\bphi,\theta)&=\mathbb{E}_{q_\theta(\bc|\bX)}\left[\|\bphi-\mathbb{E}_{p(\bz|\bc)}\left[\phi(\bz)\right]\|_2^2\right],\\
\nabla_\theta\mathcal{L}_{\mathrm{REINFORCE}}(\bX,\bphi,\theta)&=\mathbb{E}_{q_\theta(\bc|\bX)}\left[\|\bphi-\mathbb{E}_{p(\bz|\bc)}\left[\phi(\bz)\right]\|_2^2\nabla_\theta \log q_\theta(\bc|\bX)\right].
\end{aligned}
\]
As discussed in \Cref{sec:variance}, this gradient estimator can suffer from high variance because the loss for a given batch is unbounded. To mitigate this, our implementation of REINFORCE employs a baseline~\cite{mohamed_monte_2020}.
\section{Additional experimental details}\label{app:exp_details}


\subsection{Data-generating process}\label{app:dgp}
We can express our synthetic data-generating process mathematically as 
\begin{align*}
\bs{\beta}_\mu &\sim \mathrm{Uniform}([0, 1]^n)\\
\bs{\beta}_\sigma &\sim \mathrm{Uniform}([0, 1]^n)\\
\bX_i &\sim \mathrm{Normal}(\mathbf{0}, \mathbf{I}_{|\mc{E}|\times n})&\forall i\in \mc{I}\phantom{.}\\
\bc^{(s)}_{i}&\sim\mathrm{LogNormal}(\bX_i\bs{\beta}_\mu,\bX_i\bs{\beta}_\sigma)&\forall i\in \mc{I},s\in \mc{S}\phantom{.}\\
{\bz^*}^{(s)}_{i} &\in_R \mc{Z}^*(\bc^{(s)}_{i}, \bu)&\forall i\in \mc{I},s\in \mc{S}\phantom{.}\\
\bar{\mb{p}}_i&=\frac{1}{|\mc{S}|}\sum_{s\in S} {\bz^*}^{(s)}_{i} & \forall i\in \mc{I},
\end{align*}
where $\bs{\beta}_\mu$ and $\bs{\beta}_\sigma$ are global $n$-dimensional vectors. $\bX_i$ is a $|\mc{E}|\times n$ matrix, and ${\bc}^{(s)}_i$, ${\bz^*}^{(s)}_i$, and $\bar{\mb{p}}_i$ are $|\mc{E}|$-dimensional vectors. $\mc{I}$ is the set of instances and $\mc{S}$ is the set of samples for a given instance.

\subsection{Risk-averse assignment formulation}\label{app:ra_assignment}
The risk-averse assignment problem is formulated as
\begin{equation}
\begin{aligned}
    \min_{\mb{u}\in\mathbb{R}^{|\mc{K}|},v\in\mathbb{R},\mb{W}}&\quad v+\frac{1}{1-\alpha}\frac{1}{|\mc{K}|}\sum_{k\in\mc{K}} u_k\\
    \mathrm{s.t.}&\quad u_k \geq \sum_{d\in\mc{D}}\sum_{r\in\mc{R}} G^{(k)}_{dr}W_{dr} - v&\forall k\in \mc{K}\phantom{,}\\
    &\quad u_k \geq 0& \forall k\in \mc{K}\phantom{,}\\
    &\quad \sum_{\mc{D}} W_{dr} = 1&\forall r\in\mc{R}\phantom{,}\\
    &\quad \sum_{\mc{R}} W_{dr} = 1&\forall d\in\mc{D}\phantom{,}\\
    &\quad W_{dr} \in \{0,1\}&\forall d\in\mc{D},r\in\mc{R},
\end{aligned}
\end{equation}
where $\mb{u}$ is a $|\mc{K}|$-dimensional vector with $u_k$ representing the expected shortfall of scenario $k$, $v$ is the value-at-risk, $\mb{W}$ is a $|\mc{D}|\times|\mc{R}|$ assignment matrix with $W_{dr}=1$ indicating driver $d$ is assigned to rider $r$, and $\alpha$ is the risk-level (0.95 in our experiments). Given a known cost vector $\mb{g}$ and exogenous variables $\bu_{dr}\ \forall d\in\mc{D},r\in\mc{R}$, we generate a cost scenario $\mb{G}^{(k)}$ according to the following process
\begin{align*}
    \bc^{(d)} &\sim q_\theta(\bc|\bX)&\forall d\in\mc{D}\phantom{.}\\
    {\mb{z}^*}^{(dr)} &\in_R \mc{Z}^*(\bc^{(d)},\bu_{dr}) &\forall d\in \mc{D},r\in\mc{R}\phantom{.}\\
    G^{(k)}_{dr} &= \mb{g}^\top {\mb{z}^*}^{(dr)}&\forall d\in \mc{D},r\in\mc{R},\\
\end{align*}
where $G^{k}_{dr}$ is an element of the scenario cost matrix $\mb{G}^{(k)}$.
The expected solution cost under the predicted distribution is
\[
\mathbb{E}_{\zeta\sim q_\theta(\bc|\bX)}\left[g(\mb{W}^*,\bs{\zeta})\right]=\frac{1}{|\mc{K}|}\sum_{k\in\mc{K}}\sum_{d\in\mc{D}}\sum_{r\in\mc{R}}G^{(k)}_{dr}W^*_{dr}.
\]

\subsection{Baselines}\label{app:baselines}
Our PyTorch implementation including all baselines is available at {\url{https://github.com/ben-hudson/contextual-preference-distribution-learning}}.

We use the black-box hyperparameter optimization tool in Weights \& Biases to select hyperparameters for each method: beginning with 25 randomly sampled experiments, we acquire 25 more by minimizing validation loss with Bayesian optimization. Each experiment trains the given model on a randomly generated dataset with $|\mc{I}|=100$ and $|\mc{S}|=1000$, with an 80-20 train-validation split. The hyperparameter sets with the lowest validation loss are shown in \Cref{tab:hyperparams}.

\begin{table}
    \centering
    \caption{Hyperparameters used in our experiments. Perturbation scale defines the variance of the perturbing distribution; a hyperparameter in some methods and a learned quantity in others. Dashes (--) indicate not applicable to given baseline.}
    \label{tab:hyperparams}
\begin{tabular}{|r|c|c|c|c|c|}
    \hline
        \textbf{Parameter} & \textbf{AIMLE} & \textbf{DPO} & \textbf{MaxEnt IRL} & \textbf{REINFORCE} & \textbf{CPDL}\\
    \hline
        Starting LR &  1.06e-6 &5.61e-2 & 8e-4 & 3e-4 & 2.5e-3\\
        LR patience (epochs) & 20 & 15 & 5 & 7 & 15\\
        LR patience rel. tol. & 6.92\% & 0.2\% &3.64\% & 4.08\% & 3.31\%\\
        Perturbations ($K$) & 400 & 300 &--& 500 & 300\\
        Perturbation scale & 0.6 & 0.9 &--& Learned & Learned\\
        Perturbation sides & 2 & -- & -- & -- & -- \\
        Fixed-point max iters. &--&--&600&--&--\\
        Fixed-point rel. tol. &--&--&3e-8&--&--\\
        Batch size & 128 & 128 & 5 & 128 & 128\\
        Training epochs & 200 & 500 &200 &200 & 200\\
    \hline
\end{tabular}
\end{table}

\subsection{Post-decision surprise and disappointment}\label{app:disappointment}
\begin{figure}
    \centering
    \includegraphics[width=\linewidth]{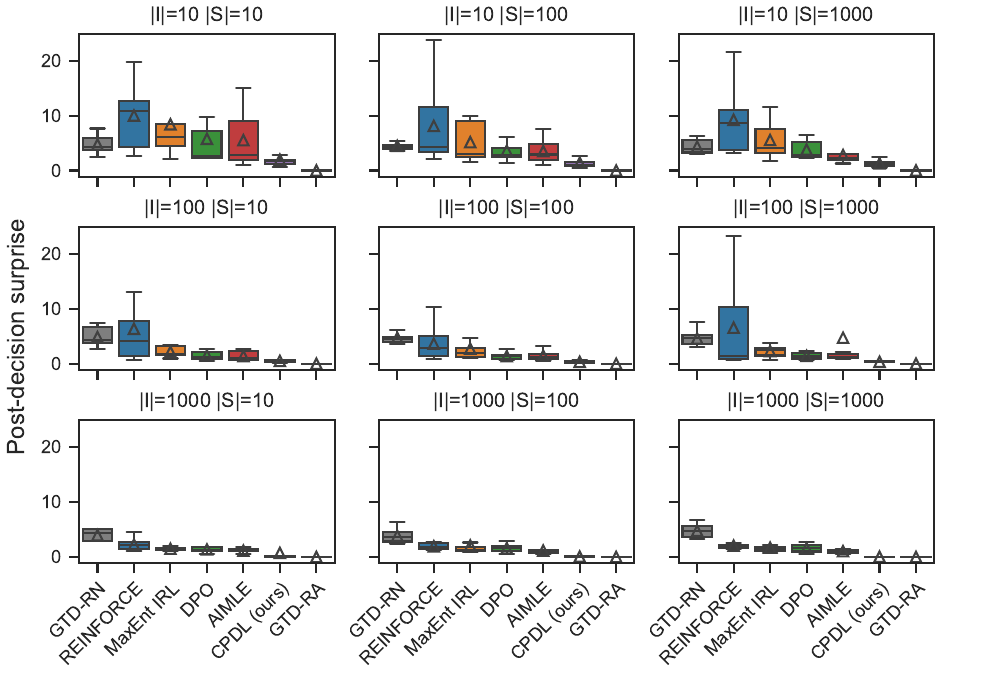}
    \includegraphics[width=\linewidth]{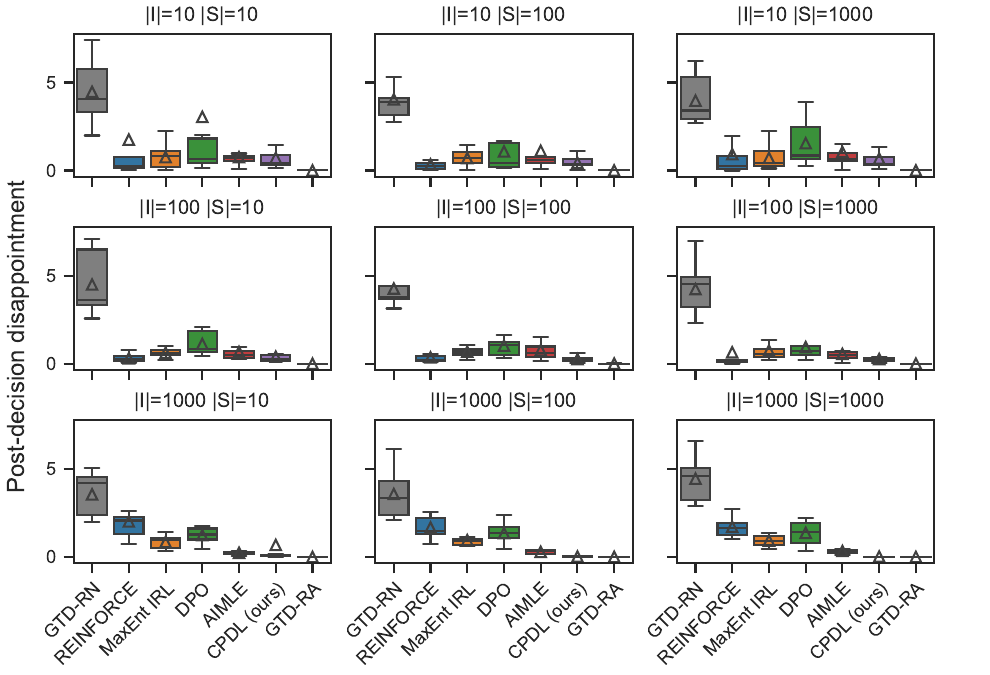}
    \caption{Post-decision surprise (top) and disappointment (bottom) as a function of training set size (rows) and observation fidelity (columns), aggregated over nine random seeds.
    While surprise penalizes early and late arrivals, disappointment only penalizes late arrivals. The combination of high surprise and low disappointment suggests overly conservative assignments.}
    \label{fig:main_results_app}
\end{figure}

Fig.~\ref{fig:main_results_app} shows post-decision surprise (computed according to \eqref{eqn:post-decision_surprise}), which penalizes both early and late arrivals, alongside post-decision disappointment, which penalizes only late arrivals. We observe that the baselines consistently achieve much lower disappointment than surprise. This suggests they make overly conservative decisions: the assignments rarely result in late drivers (low disappointment) but frequently result in early ones (high surprise).



Counterintuitively, baseline performance degrades with more data. We examined specific predictions produced by these models --- especially REINFORCE, given its similarity to our approach --- and compared them to ours. We hypothesize is that this phenomenon is a result of aspects of the data-generating process and training process. The synthetic data-generating process (App.~\ref{app:dgp}) tends to generate correlated edge costs and edge scales because the coefficients $\bs{\beta}_\mu$ and $\bs{\beta}_\sigma$ are strictly positive.
With small training sets, the model can learn the correct relationship for the distribution location, but an inverse relationship for the distribution scale (i.e., a positive $R^2$ score but a negative correlation coefficient). Consequently, the model predicts edges with high cost or high variance, when in reality they tend to have both. This results in overly conservative decisions.

While our model is susceptible to this same issue, it corrects itself with less data, which explains why its performance closely tracks GTD-RA, the risk-averse baseline with access to the ground-truth distributions. This merits further study.

\subsection{Preference distribution recovery}\label{app:dist_recovery}
\begin{figure}
    \centering
    \includegraphics[width=\linewidth]{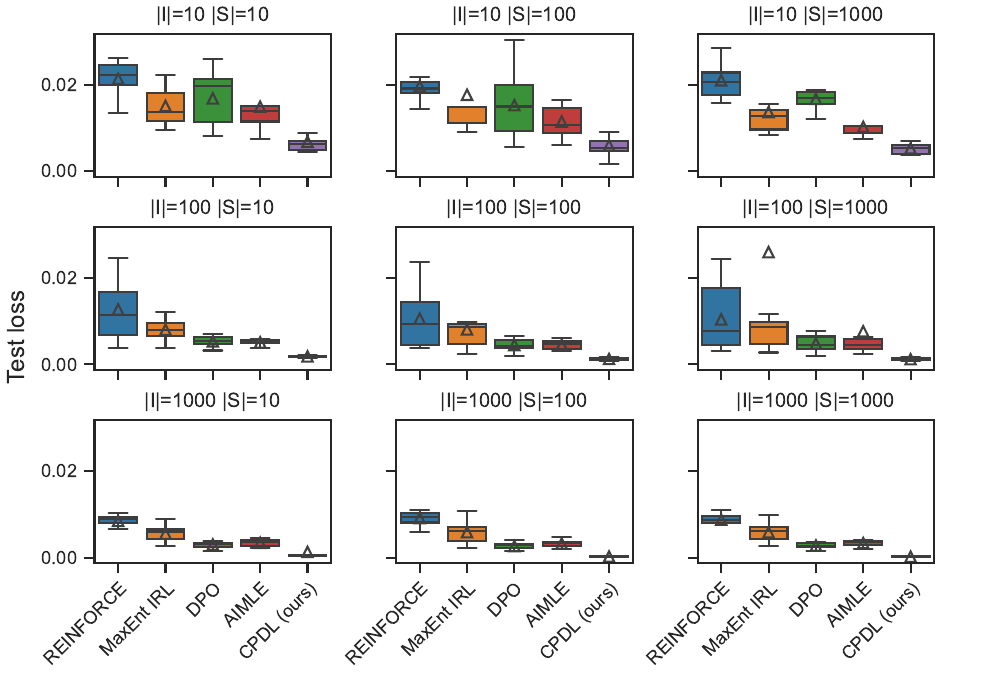}
    \caption{Test loss as a function of training set size (rows) and observation fidelity (columns), aggregated over nine random seeds.}
    \label{fig:loss_results_app}
\end{figure}

\begin{figure}
    \centering
    \includegraphics[width=\linewidth]{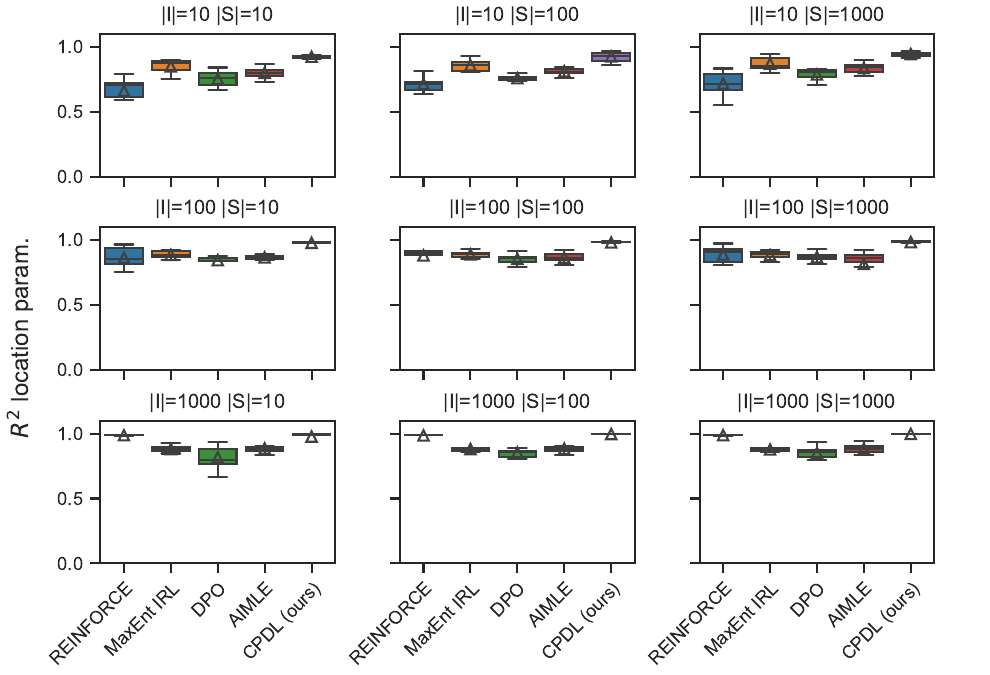}
    \includegraphics[width=\linewidth]{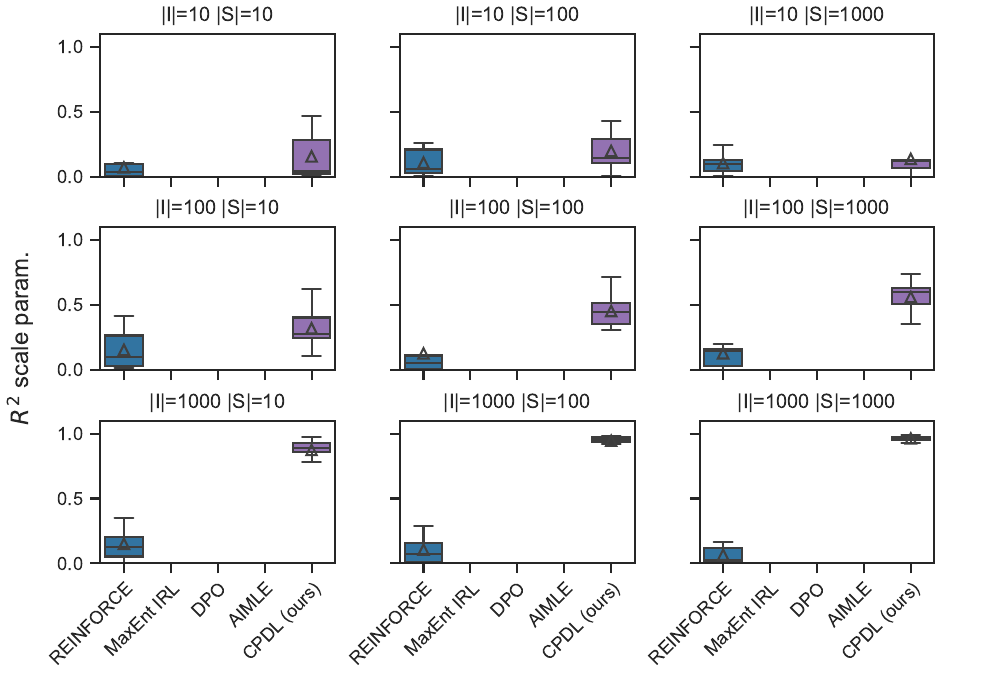}
    \caption{Recovery of the ground-truth preference distribution location (top) and scale (bottom) up to an affine transformation as a function of training set size (rows) and observation fidelity (columns), aggregated over nine random seeds. The maximum score is one and the minimum score is zero. Only REINFORCE and CPDL are able to learn parameters beyond the location.}
    \label{fig:r2_results}
\end{figure}

\Cref{fig:loss_results_app} shows test loss for additional combinations of training set size and fidelity, which measures the models' ability to reconstruct the observed choice distributions. \Cref{fig:r2_results} shows the squared correlation coefficient ($R^2$) between predicted and ground-truth preference distribution parameters, which measures their recovery up to an affine transformation.

\subsection{Matching moments beyond the mean}\label{app:higher_moments}
\begin{figure}
    \centering
    \includegraphics[width=0.99\linewidth]{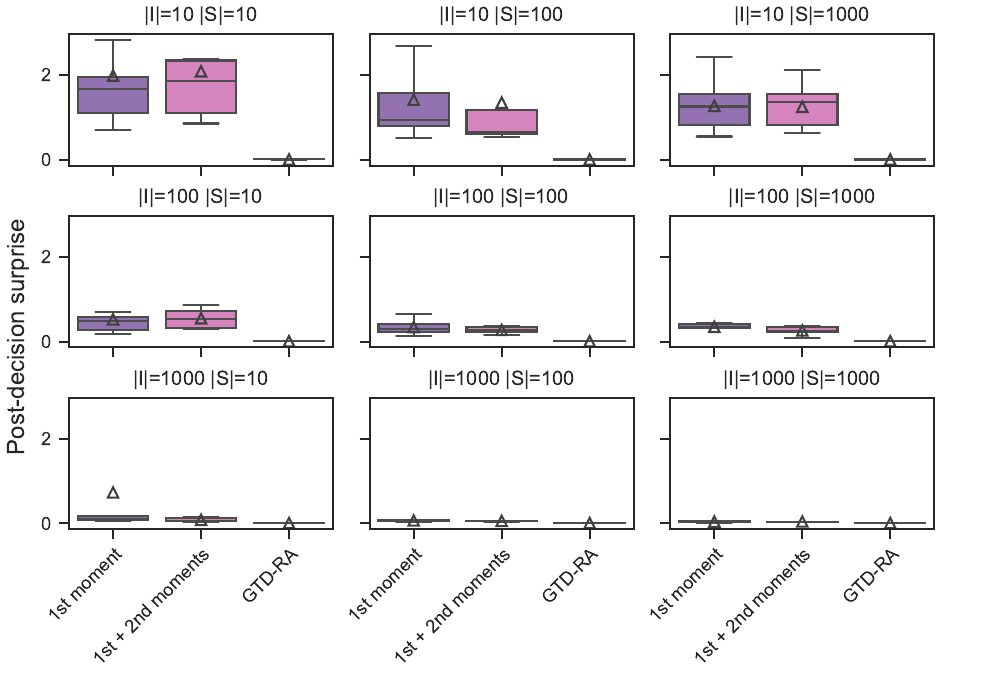}
    \includegraphics[width=0.99\linewidth]{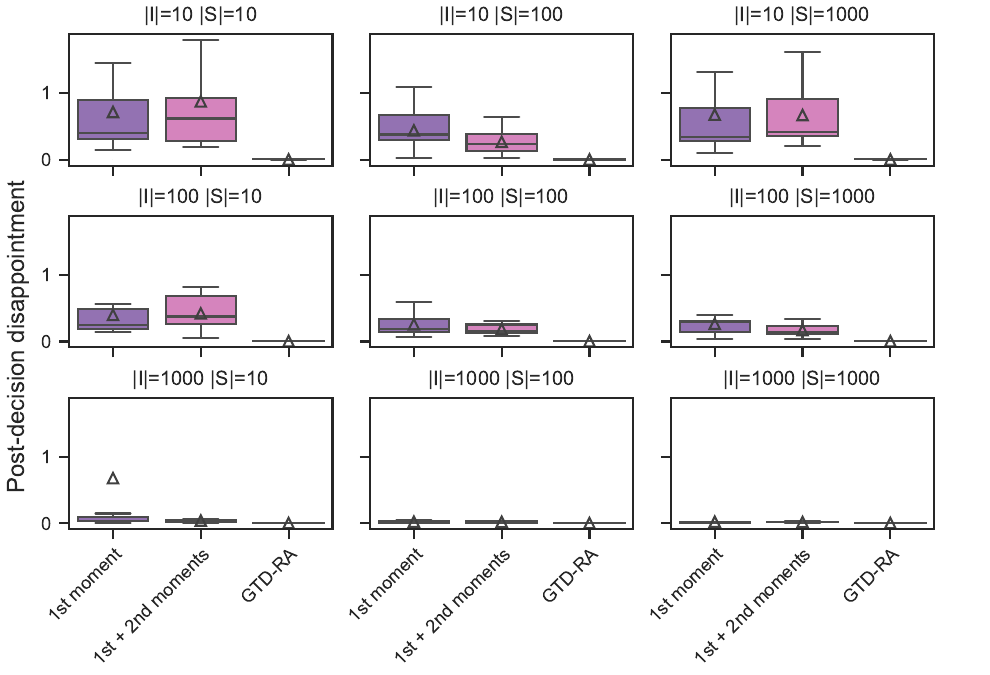}
    \caption{Post-decision surprise (top) and disappointment (bottom) for CPDL matching only the first moment compared to the first and second moments. On small (low $|\mc{I}|$) and low-fidelity (low $|\mc{S}|$) datasets, matching only the first moment tends to yield better performance. We hypothesize that the first moment is a less informative learning signal but is more robust to noise. Thus, the tradeoff shifts in favour of higher moments as dataset size and fidelity increase.}
    \label{fig:2nd_moment_matching}
\end{figure}

\Cref{fig:2nd_moment_matching} shows the impact of matching the first and second raw moments of the choice distribution, i.e. $\phi(\bz)=[\bz,\ \bz\bz^\top]$, instead of just the first.

\end{document}